\documentclass[11pt,a4paper]{article}

\usepackage[hyperref]{acl2017}

\usepackage{times}
\usepackage{latexsym}

\usepackage{url}

\usepackage{graphicx}
\usepackage{amsmath}
\usepackage{booktabs}

\aclfinalcopy 

\title{Learning Sentence Embeddings for Coherence Modelling and Beyond}

\author{Tanner Bohn \qquad Yining Hu \qquad Jinhang Zhang \qquad Charles X. Ling\\
Department of Computer Science, Western University, London, ON, Canada\\
{\tt \{tbohn,yhu534,jzha337,charles.ling\}@uwo.ca}} 

\date{}

\begin{document}
\maketitle
\begin{abstract}

We present a novel and effective technique for performing text coherence tasks while facilitating deeper insights into the data. Despite obtaining ever-increasing task performance, modern deep-learning approaches to NLP tasks often only provide users with the final network decision and no additional understanding of the data. In this work, we show that a new type of sentence embedding learned through self-supervision can be applied effectively to text coherence tasks while serving as a window through which deeper understanding of the data can be obtained. To produce these sentence embeddings, we train a recurrent neural network to take individual sentences and predict their location in a document in the form of a distribution over locations. We demonstrate that these embeddings, combined with simple visual heuristics, can be used to achieve performance competitive with state-of-the-art on multiple text coherence tasks, outperforming more complex and specialized approaches. Additionally, we demonstrate that these embeddings can provide insights useful to writers for improving writing quality and informing document structuring, and assisting readers in summarizing and locating information.

\end{abstract}

\section{Introduction}


\begin{figure}[ht]
\centering
\includegraphics[width=0.9\columnwidth]{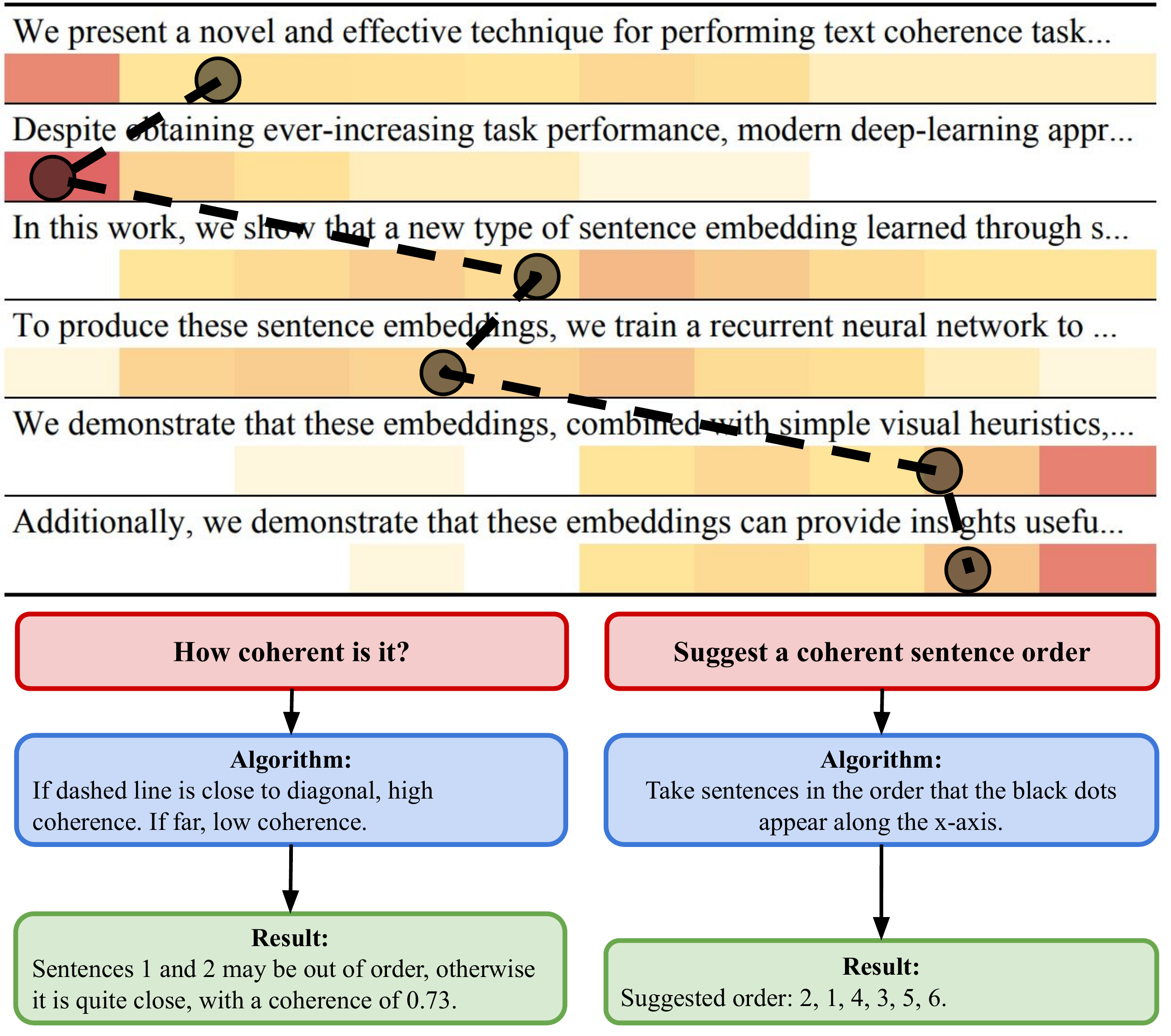}
\caption{This paper abstract is analyzed by our sentence position model trained on academic abstracts. The sentence encodings (predicted position distributions) are shown below each sentence, where white is low probability and red is high. Position quantiles are ordered from left to right. The first sentence, for example, is typical of the first sentence of abstracts as reflected in the high first-quantile value. For two text coherence tasks, we show the how the sentence encodings can easily be used to solve them. The black dots indicate the weighted average predicted position for each sentence.}
\label{fig:abstract_vis}
\end{figure}

A goal of much of NLP research is to create tools that not only assist in completing tasks, but help gain insights into the text being analyzed. This is especially true of text coherence tasks, as users are likely to wonder where efforts should be focused to improve writing or understand how text should be reorganized for improved coherence. By improving coherence, a text becomes easier to read and understand \cite{lapata2005automatic}, and in this work we particularly focus on measuring coherence in terms of sentence ordering.




Many recent approaches to NLP tasks make use of end-to-end neural approaches which exhibit ever-increasing performance, but provide little value to end-users beyond a classification or regression value~\cite{gong2016end,logeswaran2018sentence,cui2018deep}. This leaves open the question of whether we can achieve good performance on NLP tasks while simultaneously providing users with easily obtainable insights into the data. This is precisely what the work in this paper aims to do in the context of coherence analysis, by providing a tool with which users can quickly and visually gain insight into structural information about a text.
To accomplish this, we rely on the surprising importance of sentence location in many areas of natural language processing. 
If a sentence does not appear to belong where it is located, it decreases the coherence and readability of the text~\cite{lapata2005automatic}. If a sentence is located at the beginning of a document or news article, it is very likely to be a part of a high quality extractive summary~\cite{see2017get}. The location of a sentence in a scientific abstract is also an informative indicator of its rhetorical purpose~\cite{teufel1999argumentative2}. 
It thus follows that the knowledge of where a sentence should be located in a text is valuable. 

Tasks requiring knowledge of sentence position -- both relative to neighboring sentences and globally -- appear in text coherence modelling, with two important tasks being order discrimination (is a sequence of sentences in the correct order?) and sentence ordering (re-order a set of unordered sentences). Traditional methods in this area make use of manual feature engineering and established theory behind coherence~\cite{lapata2005automatic,barzilay2008modeling,grosz1995centering}. 
Modern deep-learning based approaches to these tasks tend to revolve around taking raw words and directly predicting local~\cite{li2014model,chen2016neural} or global~\cite{cui2017text,li2017neural}~ coherence scores or directly output a coherent sentence ordering~\cite{gong2016end,logeswaran2018sentence,cui2018deep}. While new deep-learning based approaches in text coherence continue to achieve ever-increasing performance, their value in real-world applications is undermined by the lack of actionable insights made available to users.


In this paper, we introduce a self-supervised approach for learning sentence embeddings which can be used effectively for text coherence tasks (Section~\ref{sec:experiments}) while also facilitating deeper understanding of the data (Section~\ref{sec:insights}). Figure~\ref{fig:abstract_vis} provides a taste of this, displaying the sentence embeddings for the abstract of this paper.  The self-supervision task we employ is that of predicting the location of a sentence in a document given only the raw text. By training a neural network on this task, it is forced to learn how the location of a sentence in a structured text is related to its syntax and semantics. As a neural model, we use a bi-directional recurrent neural network, and train it to take sentences and predict a discrete distribution over possible locations in the source text. We demonstrate the effectiveness of predicted position distributions as an accurate way to assess document coherence by performing order discrimination and sentence reordering of scientific abstracts. We also demonstrate a few types of insights that these embeddings make available to users that the predicted location of a sentence in a news article can be used to formulate an effective heuristic for extractive document summarization -- outperforming existing heuristic methods. 

The primary contributions of this work are thus:
\begin{enumerate}
    \item We propose a novel self-supervised approach to learn sentence embeddings which works by learning to map sentences to a distribution over positions in a document (Section \ref{sec:position_model}).
    \item We describe how these sentence embeddings can be applied to established coherence tasks using simple algorithms amenable to visual approximation (Section \ref{sec:model_application}).
    \item We demonstrate that these embeddings are competitive at solving text coherence tasks (Section \ref{sec:experiments}) while quickly providing access to further insights into texts (Section \ref{sec:insights}).
\end{enumerate}
\section{Predicted Position Distributions}
\label{sec:our_model}

\subsection{Overview}

By training a machine learning model to predict the location of a sentence in a body of text (conditioned upon features not trivially indicative of position), we obtain a sentence position model such that sentences predicted to be at a particular location possess properties typical of sentences found at that position. For example, if a sentence is predicted to be at the beginning of a news article, it should resemble an introductory sentence. 

In the remainder of this section we describe our neural sentence position model and then discuss how it can be applied to text coherence tasks.



\subsection{Neural Position Model}
\label{sec:position_model}

\begin{figure}[ht]
\centering
\includegraphics[width=0.8\columnwidth]{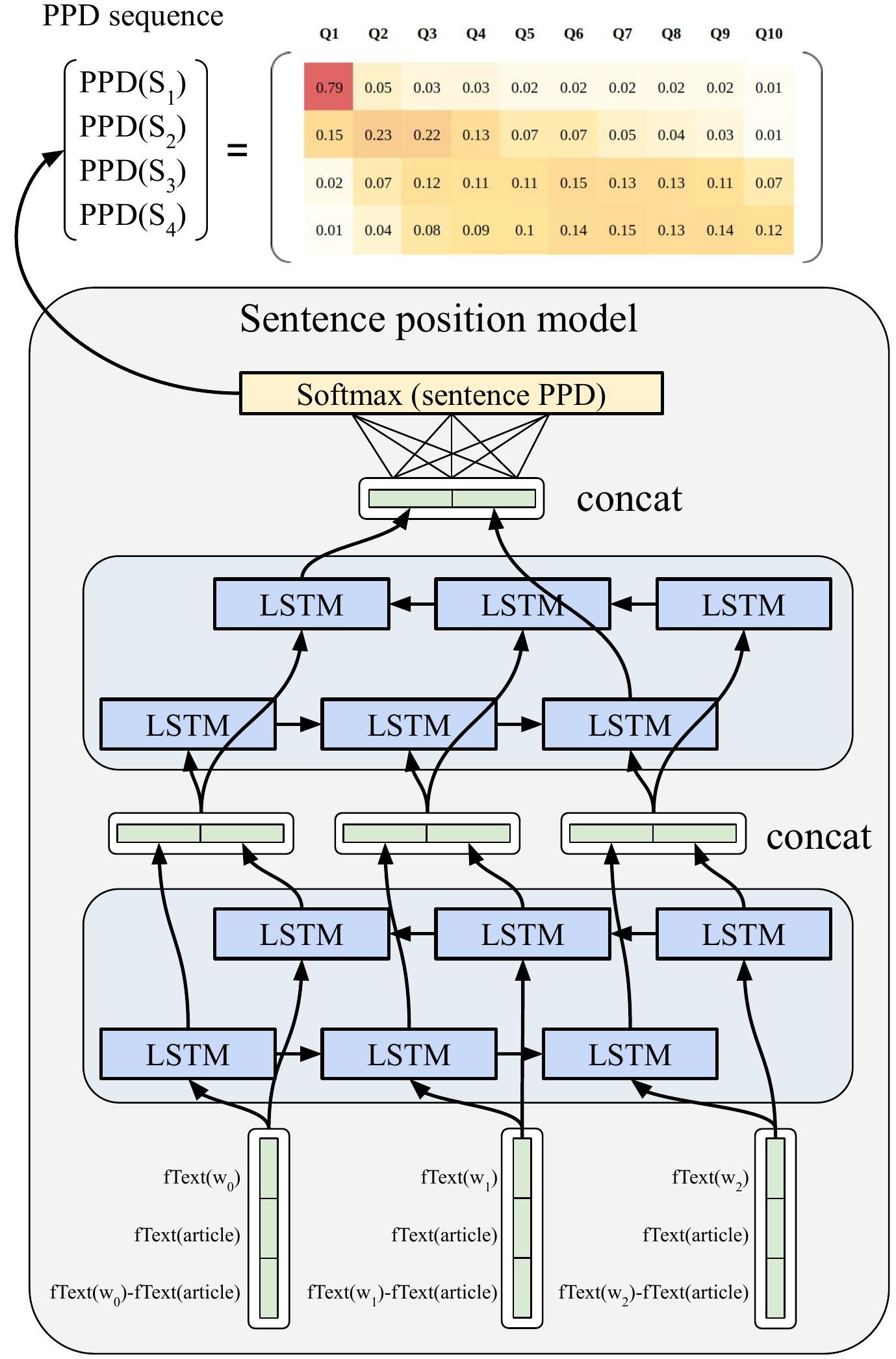}
\caption{Illustration of the sentence position model, consisting of stacked BiLSTMs. Sentences from a text are individually fed into the model to produce a PPD sequence. In this diagram we see a word sequence of length three fed into the model, which will output a single row in the PPD sequence.}
\label{fig:total_vis}
\end{figure}


The purpose of the position model is to produce sentence embeddings by predicting the position in a text of a given sentence. Training this model requires no manual labeling, needing only samples of text from the target domain. By discovering patterns in this data, the model produces sentence embeddings suitable for a variety of coherence-related NLP tasks.

\subsubsection{Model Architecture} 
To implement the position model, we use stacked bi-directional LSTMs~\cite{schuster1997bidirectional} followed by a softmax output layer. Instead of predicting a single continuous value for the position of a sentence as the fraction of the way through a document, we frame sentence position prediction as a classification problem. 

Framing the position prediction task as classification was initially motivated by the poor performance of regression models; since the task of position prediction is quite difficult, we observed that regression models would consistently make predictions very close to 0.5 (middle of the document), thus not providing much useful information. To convert the task to a classification problem, we aim to determine what quantile of the document a sentence resides in. Notationally, we will refer to the number of quantiles as $Q$. We can interpret the class probabilities behind a prediction as a discrete distribution over positions for a sentence, providing us with a predicted position distribution (PPD).
When $Q = 2$ for example, we are predicting whether a sentence is in the first or last half of a document. When $Q = 4$, we are predicting which quarter of the document it is in. In Figure~\ref{fig:total_vis} is a visualization of the neural architecture which produces PPDs of $Q = 10$.

\subsubsection{Features Used} 


The sentence position model receives an input sentence as a sequence of word encodings and outputs a single vector of dimension $Q$. Sentences are fed into the BiLSTM one at a time as a sequence of word encodings, where the encoding for each word consists of the concatenation of: (1) a pretrained word embedding, (2) the average of the pretrained word embedding for the entire document (which is constant for all words in a document), and (3) the difference of the first two components (although this information is learnable given the first two components, we found during early experimentation that it confers a small performance improvement). In addition to our own observations, the document-wide average component was also shown in~\cite{logeswaran2018sentence} to improve performance at sentence ordering, a task similar to sentence location prediction. For the pretrained word embeddings, we use 300 dimensional fastText embeddings\footnote{Available online at \url{https://fasttext.cc/docs/en/english-vectors.html}. We used the \texttt{wiki-news-300d-1M} vectors.}, shown to have excellent cross-task performance~\cite{joulin2016bag}. In Figure~\ref{fig:total_vis}, the notation $ftxt(token)$ represents converting a textual token (word or document) to its fastText embedding. The embedding for a document is the average of the embeddings for all words in it.

The features composing the sentence embeddings fed into the position model must be chosen carefully so that the order of the sentences does not directly affect the embeddings (i.e. the sentence embeddings should be the same whether the sentence ordering is permuted or not). This is because we want the predicted sentence positions to be independent of the true sentence position, and not every sentence embedding technique provides this. As a simple example, if we include the true location of a sentence in a text as a feature when training the position model, then instead of learning the connection between sentence meaning and position, the mapping would trivially exploit the known sentence position to perfectly predict the sentence quantile position. This would not allow us to observe where the sentence \textit{seems} it should be located.


\subsection{Application to Coherence Tasks}
\label{sec:model_application}

\begin{figure}[ht]
\centering
\includegraphics[width=1\columnwidth]{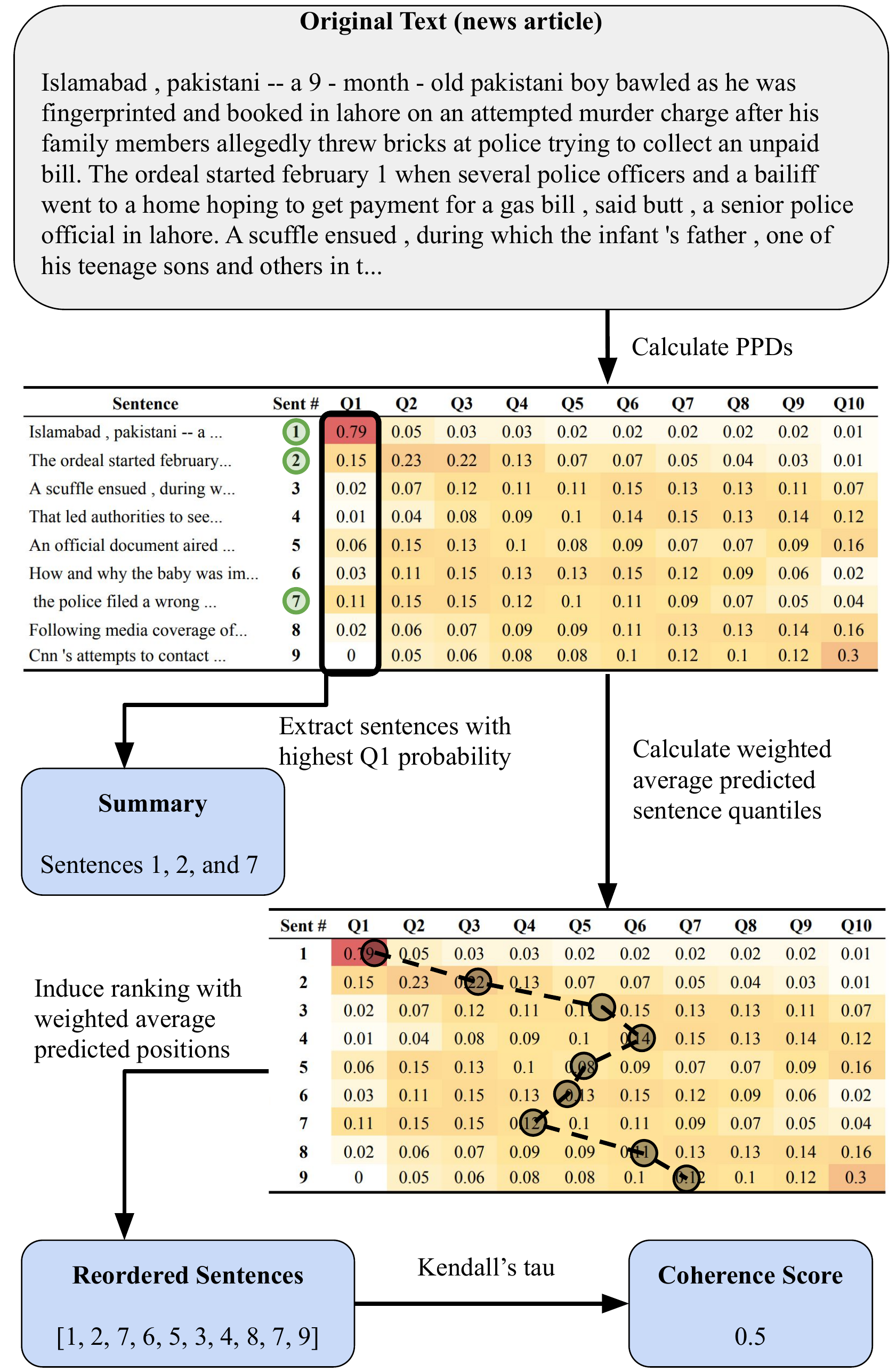}
\caption{A visualization of our NLP algorithms utilizing PPDs applied to a news article. To reorder sentences, we calculate average weighted positions (identified with black circles) to induce an ordering. Coherence is calculated with the Kendall's rank correlation coefficient between the true and induced ranking. We also show how PPDs can be used to perform summarization, as we will explore further in Section~\ref{sec:insights}.}
\label{fig:alg_vis}
\end{figure}

For the tasks of both sentence ordering and calculating coherence, PPDs can be combined with simple visually intuitive heuristics, as demonstrated in Figure~\ref{fig:alg_vis}.

\subsubsection{Sentence Ordering} 

To induce a new ordering on a sequence of sentences, $S$, we simply sort the sentence by their weighted average predicted quantile, $\hat{\mathcal{Q}}(s \in S)$, defined by:
\begin{equation}
    \hat{\mathcal{Q}}(s) = \sum_{i=1}^{Q} i \times PPD(s)_{i},
\end{equation}
where $PPD(s)$ is the $Q$-dimensional predicted position distribution/sentence embedding for the sentence $s$.

\subsubsection{Calculating coherence} 

To calculate the coherence of a text, we employ the following simple algorithm on top of the PPDs: use the Kendall's tau coefficient between the sentence ordering induced by the weighted average predicted sentence positions and the true sentence positions:
\begin{equation}
    coh = \tau((\hat{\mathcal{Q}}(s), \text{ for } s = S_{1}, ..., S_{|S|}), (1, ..., |S|)).
\end{equation}


\section{Experiments}
\label{sec:experiments}

In this section, we evaluate our PPD-based approaches on two coherence tasks and demonstrate that only minimal performance is given up by our approach to providing more insightful sentence embeddings.


\begin{table}[ht]
\centering
\resizebox{\columnwidth}{!}{%
\begin{tabular}{@{}llllll@{}}
\toprule
Task           & Dataset    & Q  & Epochs & Layer dropouts & Layer widths \\ \midrule
Order Disrcim. & Accident   & 5  & 10     & (0.4, 0.2)     & (256, 256)   \\
               & Earthquake & 10 & 5      & (0.4, 0.2)     & (256, 64)    \\
Reordering     & NeurIPS    & 15 & 20     & (0.5, 0.25)    & (256, 256)   \\ \bottomrule
\end{tabular}%
}
\caption{The neural sentence position model hyperparameters used in our coherence experiments. The following settings are used across all tasks: batch size of 32, sentence trimming/padding to a length of 25 words, the vocabulary is set to the 1000 most frequent words in the associated training set. The Adamax optimizer is used \protect\cite{kingma2014adam} with default parameters supplied by Keras \protect\cite{chollet2015keras}.}
\label{table:hyperparams}
\end{table}


\begin{table*}[ht]
\centering
\resizebox{0.7\textwidth}{!}{%
\begin{tabular}{@{}lrrlll@{}}
\toprule
                    & \multicolumn{2}{l}{\textbf{Order discrimination}}                               & \textbf{} & \multicolumn{2}{l}{\textbf{Reordering}} \\ \cmidrule(lr){2-3} \cmidrule(l){5-6} 
\textbf{Model}      & \multicolumn{1}{l}{\textbf{Accident}} & \multicolumn{1}{l}{\textbf{Earthquake}} &           & \textbf{Acc}      & \textbf{$\tau$}     \\ \midrule
Random              & 50                                    & 50                                      &           & 15.6              & 0                   \\
Entiry Grid         & 90.4                                  & 87.2                                    &           & 20.1              & 0.09                \\
Window network      & \multicolumn{1}{c}{-}                 & \multicolumn{1}{c}{-}                   &           & 41.7              & 0.59                \\
LSTM\_PtrNet        & 93.7                                  & 99.5                                    &           & 50.9              & 0.67                \\
RNN Decoder         & \multicolumn{1}{c}{-}                 & \multicolumn{1}{c}{-}                   &           & 48.2              & 0.67                \\
Varient-LSTM+PtrNet & 94.4                                  & 99.7                                    &           & 51.6              & \textbf{0.72}       \\
ATTOrderNet         & \textbf{96.2}                         & \textbf{99.8}                           &           & \textbf{56.1}     & \textbf{0.72}       \\ \midrule
PPDs                & 94.4                                  & 99.3                                    &           & 54.9              & \textbf{0.72}       \\ \bottomrule
\end{tabular}%
}
\caption{Results on the order discrimination and sentence reordering coherence tasks. Our approach trades only a small decrease in performance for improved utility of the sentence embeddings over other approaches, achieving close to or the same as the state-of-the-art.}
\label{table:coherence_results}
\end{table*}

\textbf{Order discrimination setup.} For order discrimination, we use the Accidents and Earthquakes datasets from~\cite{barzilay2008modeling} which consists of aviation accident reports and news articles related to earthquakes respectively. The task is to determine which of a permuted ordering of the sentences and the original ordering is the most coherent (in the original order), for twenty such permutations. Since these datasets only contain training and testing partitions, we follow~\cite{li2014model} and perform 10-fold cross-validation for hyperparameter tuning. Performance is measured with the accuracy with which the permuted sentences are identified. For example, the Entity Grid baseline in Table \ref{table:coherence_results} gets 90.4\% accuracy because given a shuffled report and original report, it correctly classifies them 90.4\% of the time. 

\textbf{Sentence ordering setup.} For sentence ordering, we use past NeurIPS abstracts to compare with previous works. While our validation and test partitions are nearly identical to those from~\cite{logeswaran2018sentence}, we use a publicly available dataset\footnote{\url{https://www.kaggle.com/benhamner/nips-papers}} which is missing the years 2005, 2006, and 2007 from the training set (\cite{logeswaran2018sentence} collected data from 2005 - 2013). Abstracts from 2014 are used for validation, and 2015 is used for testing. To measure performance, we report both reordered sentence position accuracy as well as Kendall's rank correlation coefficient. For example, the Random baseline correctly predicts the index of sentences 15.6\% of the time, but there is no correlation between the predicted ordering and true ordering, so $\tau = 0$.

\textbf{Training and tuning.} Hyperparameter tuning for both tasks is done with a random search, choosing the hyperparameter set with the best validation score averaged across the 10 folds for order discrimination dataset and for three trials for the sentence reordering task. The final hyperparameters chosen are in Table~\ref{table:hyperparams}.

\textbf{Baselines.} We compare our results against a random baseline, the traditional Entity Grid approach from~\cite{barzilay2008modeling}, Window network~\cite{li2014model}, LSTM+PtrNet~\cite{gong2016end}, RNN Decoder and Varient-LSTM+PtrNet from~\cite{logeswaran2018sentence}, and the most recent state-of-the art ATTOrderNet~\cite{cui2018deep}.

\textbf{Results.} Results for both coherence tasks are collected in Table~\ref{table:coherence_results}. 
For the order discrimination task, we find that on both datasets, our PPD-based approach only slightly underperforms ATTOrderNet \cite{cui2018deep}, with performance similar to the LSTM+PtrNet approaches \cite{gong2016end,logeswaran2018sentence}.
On the more difficult sentence reordering task, our approach exhibits performance closer to the state-of-the-art, achieving the same ranking correlation and only slightly lower positional accuracy. Given that the publicly available training set for the reordering task is slightly smaller than that used in previous work, it is possible that more data would allow our approach to achieve even better performance. In the next section we will discuss the real-world value offered by our approach that is largely missing from existing approaches.

\section{Actionable Insights}
\label{sec:insights}

\begin{figure*}[ht]
	\centering
	\includegraphics[width=1\textwidth]{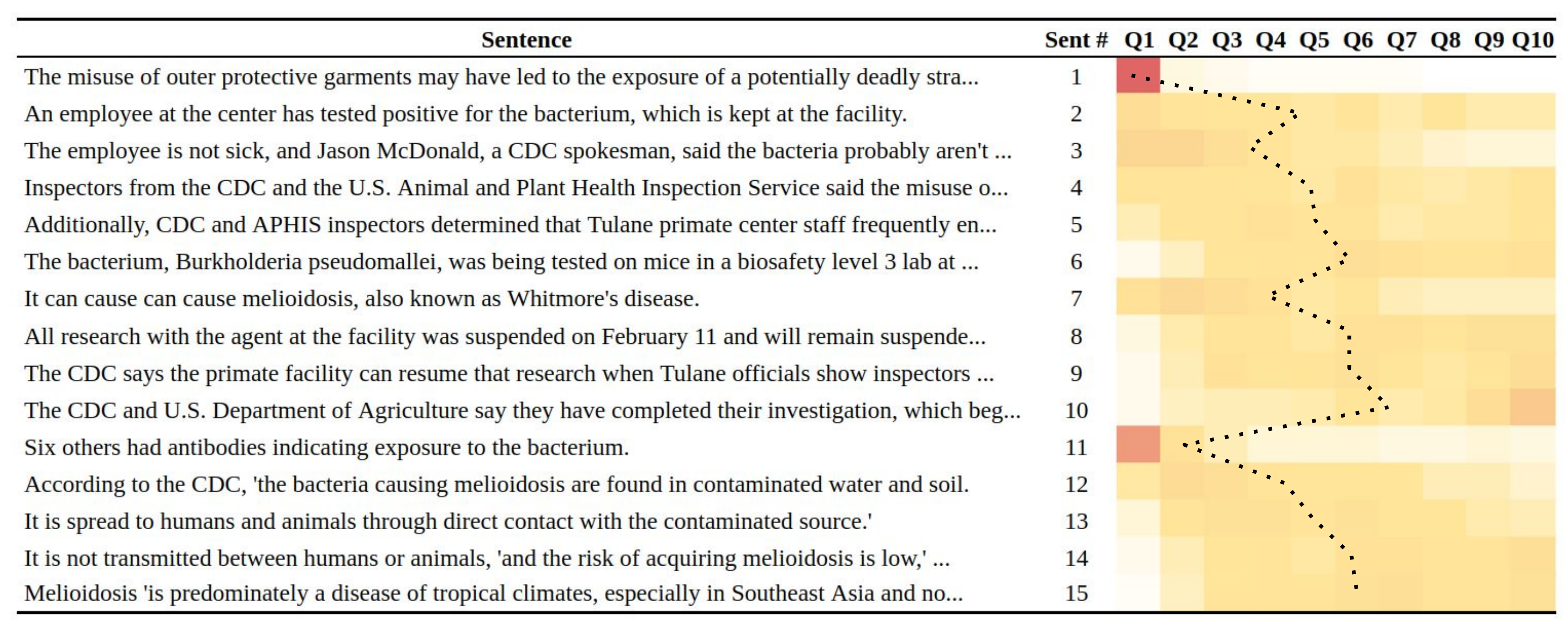}
	\caption{The PPDs for a CNN article. (full text available at \url{http://web.archive.org/web/20150801040019id\_/http://www.cnn.com/2015/03/13/us/tulane-bacteria-exposure/}). The dashed line shows the weighted average predicted sentence positions.}
	\label{fig:insight_ex}
\end{figure*}

A primary benefit of applying PPDs to coherence-related tasks is the ability to gain deeper insights into the data. In this section, we will demonstrate the following in particular: (1) how PPDs can quickly be used to understand how the coherence of a text may be improved, (2) how the existence of multiple coherence subsections may be identified, and (3) how PPDs can allow users to locate specific types of information without reading a single word, a specific case of which is extractive summarization. For demonstrations, we will use the news article presented in Figure~\ref{fig:insight_ex}.

\subsection{Improving Coherence}

	For a writer to improve their work, understanding the incoherence present is important. Observing the PPD sequence for the article in Figure~\ref{fig:insight_ex} makes it easy to spot areas of potential incoherence: they occur where consecutive PPDs are significantly different (from sentences 1 to 2, 6 to 7, and 10 to 11). In this case, the writer may determine that sentence 2 is perhaps not as introductory as it should be. The predicted incoherence between sentences 10 and 11 is more interesting, and as we will see next, the writer may realize that this incoherence may be okay to retain.


\subsection{Identifying Subsections}

	In Figure~\ref{fig:insight_ex}, we see rough progressions of introductory-type sentences to conclusory-type sentences between sentences 1 and 10 and sentences 11 and 15. This may indicate that the article is actually composed of two coherent subsections, which means that the incoherence between sentences 10 and 11 is expected and natural. By being able to understand where subsections may occur in a document, a writer can make informed decisions on where to split a long text into more coherent chunks or paragraphs. Knowing where approximate borders between ideas in a document exist may also help readers skim the document to find desired information more quickly, as further discussed in the next subsection.


\subsection{Locating Information and Summarization}

    \begin{table*}[ht]
\centering
\resizebox{\textwidth}{!}{%
\begin{tabular}{lrrr}
\hline
\textbf{Model (lead baseline source)} & \multicolumn{1}{c}{\textbf{ROUGE-1}} & \multicolumn{1}{c}{\textbf{ROUGE-2}} & \multicolumn{1}{c}{\textbf{ROUGE-L}} \\ \hline
Lead-3 \cite{nallapati2017summarunner} & 39.2 & 15.7 & 35.5 \\
Lead-3 \cite{see2017get} & 40.3 & 17.7 & 36.6 \\
Lead-3 (Ours) & 35.8 & 15.9 & 33.5 \\ \hline
SummaRuNNer \cite{nallapati2017summarunner} (\cite{nallapati2017summarunner}) & 39.6 & 16.2 & 35.3 \\
Pointer-generator \cite{see2017get} (\cite{see2017get}) & 39.5 & 17.3 & 36.4 \\
RL \cite{paulus2017deep} (\cite{nallapati2017summarunner}) & 41.2 & 15.8 & 39.1 \\ \hline
TextRank \cite{mihalcea2004textrank} (ours) & 26.2 & 11.1 & 24.3 \\
Luhn \cite{luhn1958automatic} (ours) & 26.4 & 11.2 & 24.5 \\
SumBasic \cite{nenkova2005impact} (ours) & 27.8 & 10.4 & 26.0 \\
LexRank \cite{erkan2004lexrank} (ours) & 28.4 & 11.6 & 26.3 \\
PPDs (ours) & \textbf{30.1} & \textbf{12.6} & \textbf{28.2} \\ \hline
\end{tabular}%
}
\caption{ROUGE scores on the CNN/DailyMail summarization task. Our PPD-based heuristic outperforms the suite of established heuristic summarizers. However, the higher performance of the deep-learning models demonstrates that training explicitly for summarization is beneficial.}
\label{table:summarization_results}
\end{table*}


	When reading a new article, readers well-versed in the subject of the article may want to skip high-level introductory comments and jump straight to the details. For those unfamiliar with the content or triaging many articles, this introductory information is important to determine the subject matter. Using PPDs, locating these types of information quickly should be easy for readers, even when the document has multiple potential subsections. In Figure~\ref{fig:insight_ex}, sentences 1 and 11 likely contain introductory information (since the probability of occurring in the first quantiles is highest), the most conclusory-type information is in sentence 10, and lower-level details are likely spread among the remaining sentences.

	Locating sentences with the high-level details of a document is reminiscent of the task of extractive summarization, where significant research has been performed~\cite{nenkova2011automatic,nenkova2012survey}. It is thus natural to ask how well a simple PPD-based approach performs at summarization. To answer this question, the summarization algorithm we will use is: select the $n$ sentences with the highest $PPD(s \in S)_0$ value, where $S$ is the article being extractively summarized down to $n$ sentences. For the article in Figure~\ref{fig:insight_ex}, sentences 1, 11, and 3 would be chosen since they have the highest first-quantile probabilities. This heuristic is conceptually similar to the Lead heuristic, where sentences that actually occur at the start of the document are chosen to be in the summary. Despite its simplicity, the Lead heuristic often achieves near state-of-the-art results~\cite{see2017get}.



	We experiment on the non-anonymized CNN/DailyMail dataset~\cite{hermann2015teaching} and evaluate with full-length ROUGE-1, -2, and -L F1 scores~\cite{lin2003automatic}.
	For the neural position model, we choose four promising sets of hyperparameters identified during the hyperparameter search for the sentence ordering task in Section~\ref{sec:experiments} and train each sentence position model on 10K of the 277K training articles (which provides our sentence position model with over 270K sentences to train on). Test results are reported for the model with the highest validation score. The final hyperparameters chosen for this sentence location model are: $Q$ = 10, epochs = 10, layer dropouts = (0.4, 0.2), layer widths = (512, 64).
	
	We compare our PPD-based approach to other heuristic approaches\footnote{Implementations provided by Sumy library, available at \url{https://pypi.python.org/pypi/sumy}.}. For completeness, we also include results of deep-learning based approaches and their associated Lead baselines evaluated using full-length ROUGE scores on the non-anonymized CNN/DailyMail dataset.
	

    Table~\ref{table:summarization_results} contains the the comparison between our PPD-based summarizer and several established heuristic summarizers. We observe that our model has ROUGE scores superior to the other heuristic approaches by a margin of approximately 2 points for ROUGE-1 and -L and 1 point for ROUGE-2. In contrast, the deep-learning approaches trained explicitly for summarization achieve even higher scores, suggesting that there is more to a good summary than the sentences simply being introductory-like.

\section{Related Work}
\label{sec:related}


Extensive research has been done on text coherence, motivated by downstream utility of coherence models. In addition to the applications we demonstrate in Section~\ref{sec:insights}, established applications include determining the readability of a text (coherent texts are easier to read)~\cite{barzilay2008modeling}, refinement of multi-document summaries~\cite{barzilay2002inferring}, and essay scoring~\cite{farag2018neural}.

Traditional methods to coherence modelling utilize established theory and handcrafted linguistic features \cite{grosz1995centering,lapata2003probabilistic}. 
The Entity Grid model~\cite{lapata2005automatic,barzilay2008modeling} is an influential traditional approach which works by first constructing a sentence~$\times$~discourse entities (noun phrases) occurrence matrix, keeping track of the syntactic role of each entity in each sentence. Sentence transition probabilities are then calculated using this representation and used as a feature vector as input to a SVM classifier trained to rank sentences on coherence.

Newer methods utilizing neural networks and deep learning can be grouped together by whether they indirectly or directly produce an ordering given an unordered set of sentences.

\textbf{Indirect ordering.} Approaches in the indirect case include Window network~\cite{li2014model}, Pairwise Ranking Model~\cite{chen2016neural}, the deep coherence model from~\cite{cui2017text}, and the discriminative model from~\cite{li2017neural}. These approaches are trained to take a set of sentences (anywhere from two~\cite{chen2016neural} or three~\cite{li2014model} to the whole text~\cite{cui2017text,li2017neural}) and predict whether the component sentences are already in a coherent order. A final ordering of sentences is constructed by maximizing coherence of sentence subsequences.

\textbf{Direct ordering.} Approaches in the direct case include~\cite{gong2016end,logeswaran2018sentence,cui2018deep}. These model are trained to take a set of sentences, encode them using some technique, and with a recurrent neural network decoder, output the order in which the sentences would coherently occur.

Models in these two groups all use similar high-level architectures: a recurrent or convolutional sentence encoder, an optional paragraph encoder, and then either predicting coherence from that encoding or iteratively reconstructing the ordering of the sentences. The PPD-based approaches described in Section~\ref{sec:our_model} take a novel route of directly predicting location information of each sentence. Our approaches are thus similar to the direct approaches in that position information is directly obtained (here, in the PPDs), however the position information produced by our model is much more rich than simply the index of the sentence in the new ordering. With the set of indirect ordering approaches, our model approach to coherence modelling shares the property that induction of an ordering upon the sentences is only done after examining all of the sentence embeddings and explicitly arranging them in the most coherent fashion.











\section{Conclusions}
\label{sec:conclusions}

The ability to facilitate deeper understanding of texts is an important, but recently ignored, property for coherence modelling approaches. In an effort to improve this situation, we present a self-supervised approach to learning sentence embeddings, which we call PPDs, that rely on the connection between the meaning of a sentence and its location in a text. We implement the new sentence embedding technique with a recurrent neural network trained to map a sentence to a discrete distribution indicating where in the text the sentence is likely located. These PPDs have the useful property that a high probability in a given quantile indicates that the sentence is typical of sentences that would occur at the corresponding location in the text.

We demonstrate how these PPDs can be applied to coherence tasks with algorithms simple enough such that they can be visually performed by users while achieving near state-of-the-art, outperforming more complex and specialized systems. We also demonstrate how PPDs can be used to obtain various insights into data, including how to go about improving the writing, how to identify potential subsections, and how to locate specific types of information, such as introductory or summary information. As a proof-of-concept, we additionally show that despite PPDs not being designed for the task, they can be used to create a heuristic summarizer which outperforms comparable heuristic summarizers.


In future work, it would be valuable to evaluate our approach on texts from a wider array of domains and with different sources of incoherence. In particular, examining raw texts identified by humans as lacking coherence could be performed, to determine how well our model correlates with human judgment. Exploring how the algorithms utilizing PPDs may be refined for improved performance on the wide variety of coherence-related tasks may also prove fruitful. We are also interested in examining how PPDs may assist with other NLP tasks such as text generation or author identification. 



\section*{Acknowledgments}

We acknowledge the support of the Natural Sciences and Engineering Research Council of Canada (NSERC) through the Discovery Grants Program. NSERC invests annually over \$1 billion in people, discovery and innovation.

%
%
%
\bibliographystyle{acl_natbib}
%
\bibliography{mybib}

\begin{thebibliography}{}
\expandafter\ifx\csname natexlab\endcsname\relax\def\natexlab#1{#1}\fi

\bibitem[{Barzilay and Elhadad(2002)}]{barzilay2002inferring}
Regina Barzilay and Noemie Elhadad. 2002.
\newblock Inferring strategies for sentence ordering in multidocument news
  summarization.
\newblock {\em Journal of Artificial Intelligence Research\/} .

\bibitem[{Barzilay and Lapata(2008)}]{barzilay2008modeling}
Regina Barzilay and Mirella Lapata. 2008.
\newblock Modeling local coherence: An entity-based approach.
\newblock {\em Computational Linguistics\/} 34(1):1--34.

\bibitem[{Chen et~al.(2016)Chen, Qiu, and Huang}]{chen2016neural}
Xinchi Chen, Xipeng Qiu, and Xuanjing Huang. 2016.
\newblock Neural sentence ordering.
\newblock {\em arXiv preprint arXiv:1607.06952\/} .

\bibitem[{Chollet et~al.(2015)}]{chollet2015keras}
Fran\c{c}ois Chollet et~al. 2015.
\newblock Keras.
\newblock \url{https://github.com/keras-team/keras}.

\bibitem[{Cui et~al.(2018)Cui, Li, Chen, and Zhang}]{cui2018deep}
Baiyun Cui, Yingming Li, Ming Chen, and Zhongfei Zhang. 2018.
\newblock \href{http://aclweb.org/anthology/D18-1465}{Deep attentive sentence
  ordering network}.
\newblock In {\em Proceedings of the 2018 Conference on Empirical Methods in
  Natural Language Processing\/}. Association for Computational Linguistics,
  pages 4340--4349.
\newblock
  \href{http://aclweb.org/anthology/D18-1465}{http://aclweb.org/anthology/D18-1465}.

\bibitem[{Cui et~al.(2017)Cui, Li, Zhang, and Zhang}]{cui2017text}
Baiyun Cui, Yingming Li, Yaqing Zhang, and Zhongfei Zhang. 2017.
\newblock Text coherence analysis based on deep neural network.
\newblock In {\em Proceedings of the 2017 ACM on Conference on Information and
  Knowledge Management\/}. ACM, pages 2027--2030.

\bibitem[{Erkan and Radev(2004)}]{erkan2004lexrank}
G{\"u}nes Erkan and Dragomir~R Radev. 2004.
\newblock Lexrank: Graph-based lexical centrality as salience in text
  summarization.
\newblock {\em Journal of Artificial Intelligence Research\/} 22:457--479.

\bibitem[{Farag et~al.(2018)Farag, Yannakoudakis, and
  Briscoe}]{farag2018neural}
Youmna Farag, Helen Yannakoudakis, and Ted Briscoe. 2018.
\newblock Neural automated essay scoring and coherence modeling for
  adversarially crafted input.
\newblock {\em arXiv preprint arXiv:1804.06898\/} .

\bibitem[{Gong et~al.(2016)Gong, Chen, Qiu, and Huang}]{gong2016end}
Jingjing Gong, Xinchi Chen, Xipeng Qiu, and Xuanjing Huang. 2016.
\newblock End-to-end neural sentence ordering using pointer network.
\newblock {\em arXiv preprint arXiv:1611.04953\/} .

\bibitem[{Grosz et~al.(1995)Grosz, Weinstein, and Joshi}]{grosz1995centering}
Barbara~J Grosz, Scott Weinstein, and Aravind~K Joshi. 1995.
\newblock Centering: A framework for modeling the local coherence of discourse.
\newblock {\em Computational linguistics\/} 21(2):203--225.

\bibitem[{Hermann et~al.(2015)Hermann, Kocisky, Grefenstette, Espeholt, Kay,
  Suleyman, and Blunsom}]{hermann2015teaching}
Karl~Moritz Hermann, Tomas Kocisky, Edward Grefenstette, Lasse Espeholt, Will
  Kay, Mustafa Suleyman, and Phil Blunsom. 2015.
\newblock Teaching machines to read and comprehend.
\newblock In {\em Advances in Neural Information Processing Systems\/}. pages
  1693--1701.

\bibitem[{Joulin et~al.(2016)Joulin, Grave, Bojanowski, and
  Mikolov}]{joulin2016bag}
Armand Joulin, Edouard Grave, Piotr Bojanowski, and Tomas Mikolov. 2016.
\newblock Bag of tricks for efficient text classification.
\newblock {\em arXiv preprint arXiv:1607.01759\/} .

\bibitem[{Kingma and Ba(2014)}]{kingma2014adam}
Diederik~P Kingma and Jimmy Ba. 2014.
\newblock Adam: A method for stochastic optimization.
\newblock {\em arXiv preprint arXiv:1412.6980\/} .

\bibitem[{Lapata(2003)}]{lapata2003probabilistic}
Mirella Lapata. 2003.
\newblock Probabilistic text structuring: Experiments with sentence ordering.
\newblock In {\em Proceedings of the 41st Annual Meeting on Association for
  Computational Linguistics-Volume 1\/}. Association for Computational
  Linguistics, pages 545--552.

\bibitem[{Lapata and Barzilay(2005)}]{lapata2005automatic}
Mirella Lapata and Regina Barzilay. 2005.
\newblock Automatic evaluation of text coherence: Models and representations.
\newblock In {\em IJCAI\/}. volume~5, pages 1085--1090.

\bibitem[{Li and Hovy(2014)}]{li2014model}
Jiwei Li and Eduard Hovy. 2014.
\newblock A model of coherence based on distributed sentence representation.
\newblock In {\em Proceedings of the 2014 Conference on Empirical Methods in
  Natural Language Processing (EMNLP)\/}. pages 2039--2048.

\bibitem[{Li and Jurafsky(2017)}]{li2017neural}
Jiwei Li and Dan Jurafsky. 2017.
\newblock Neural net models of open-domain discourse coherence.
\newblock In {\em Proceedings of the 2017 Conference on Empirical Methods in
  Natural Language Processing\/}. pages 198--209.

\bibitem[{Lin and Hovy(2003)}]{lin2003automatic}
Chin-Yew Lin and Eduard Hovy. 2003.
\newblock Automatic evaluation of summaries using n-gram co-occurrence
  statistics.
\newblock In {\em Proceedings of the 2003 Conference of the North American
  Chapter of the Association for Computational Linguistics on Human Language
  Technology-Volume 1\/}. Association for Computational Linguistics, pages
  71--78.

\bibitem[{Logeswaran et~al.(2018)Logeswaran, Lee, and
  Radev}]{logeswaran2018sentence}
Lajanugen Logeswaran, Honglak Lee, and Dragomir Radev. 2018.
\newblock Sentence ordering and coherence modeling using recurrent neural
  networks.
\newblock In {\em Thirty-Second AAAI Conference on Artificial Intelligence\/}.

\bibitem[{Luhn(1958)}]{luhn1958automatic}
Hans~Peter Luhn. 1958.
\newblock The automatic creation of literature abstracts.
\newblock {\em IBM Journal of research and development\/} 2(2):159--165.

\bibitem[{Mihalcea and Tarau(2004)}]{mihalcea2004textrank}
Rada Mihalcea and Paul Tarau. 2004.
\newblock Textrank: Bringing order into text.
\newblock In {\em Proceedings of the 2004 conference on empirical methods in
  natural language processing\/}.

\bibitem[{Nallapati et~al.(2017)Nallapati, Zhai, and
  Zhou}]{nallapati2017summarunner}
Ramesh Nallapati, Feifei Zhai, and Bowen Zhou. 2017.
\newblock Summarunner: A recurrent neural network based sequence model for
  extractive summarization of documents.
\newblock In {\em AAAI\/}. pages 3075--3081.

\bibitem[{Nenkova and McKeown(2012)}]{nenkova2012survey}
Ani Nenkova and Kathleen McKeown. 2012.
\newblock A survey of text summarization techniques.
\newblock In {\em Mining text data\/}, Springer, pages 43--76.

\bibitem[{Nenkova et~al.(2011)Nenkova, McKeown et~al.}]{nenkova2011automatic}
Ani Nenkova, Kathleen McKeown, et~al. 2011.
\newblock Automatic summarization.
\newblock {\em Foundations and Trends{\textregistered} in Information
  Retrieval\/} 5(2--3):103--233.

\bibitem[{Nenkova and Vanderwende(2005)}]{nenkova2005impact}
Ani Nenkova and Lucy Vanderwende. 2005.
\newblock The impact of frequency on summarization.
\newblock {\em Microsoft Research, Redmond, Washington, Tech. Rep.
  MSR-TR-2005\/} 101.

\bibitem[{Paulus et~al.(2017)Paulus, Xiong, and Socher}]{paulus2017deep}
Romain Paulus, Caiming Xiong, and Richard Socher. 2017.
\newblock A deep reinforced model for abstractive summarization.
\newblock {\em arXiv preprint arXiv:1705.04304\/} .

\bibitem[{Schuster and Paliwal(1997)}]{schuster1997bidirectional}
Mike Schuster and Kuldip~K Paliwal. 1997.
\newblock Bidirectional recurrent neural networks.
\newblock {\em IEEE Transactions on Signal Processing\/} 45(11):2673--2681.

\bibitem[{See et~al.(2017)See, Liu, and Manning}]{see2017get}
Abigail See, Peter~J Liu, and Christopher~D Manning. 2017.
\newblock Get to the point: Summarization with pointer-generator networks.
\newblock In {\em Proceedings of the 55th Annual Meeting of the Association for
  Computational Linguistics (Volume 1: Long Papers)\/}. volume~1, pages
  1073--1083.

\bibitem[{Teufel et~al.(1999)}]{teufel1999argumentative2}
Simone Teufel et~al. 1999.
\newblock {\em Argumentative zoning: Information extraction from scientific
  text\/}.
\newblock Ph.D. thesis, Citeseer.

\end{thebibliography}

\end{document}